\documentclass[letterpaper]{article} 
\usepackage{aaai25}  
\usepackage{times}  
\usepackage{helvet}  
\usepackage{courier}  
\usepackage[hyphens]{url}  
\usepackage{graphicx} 
\usepackage{natbib}  
\usepackage{caption} 
\usepackage{algorithm}
\usepackage{algorithmic}
\usepackage{multirow}
\usepackage{makecell}
\usepackage{amsmath}
\usepackage{amsfonts}
\usepackage{amssymb}
\usepackage{comment}
\usepackage{bm}
\usepackage{pifont}
\usepackage{subcaption}

\usepackage{newfloat}
\usepackage{listings}
\DeclareCaptionStyle{ruled}{labelfont=normalfont,labelsep=colon,strut=off} 
\floatstyle{ruled}
\newfloat{listing}{tb}{lst}{}
\floatname{listing}{Listing}
\title{Maximizing the Position Embedding for Vision Transformers \\ with Global Average Pooling}
\author{
    Wonjun Lee\textsuperscript{\rm 1,\rm 2}, 
    Bumsub Ham\textsuperscript{\rm 1}, 
    Suhyun Kim\textsuperscript{\rm 2}\equalcontrib
}
\affiliations{
    \textsuperscript{\rm 1}Yonsei University, Republic of Korea \\
    \textsuperscript{\rm 2}Korea Institute of Science and Technology, Republic of Korea \\
    \{velpegor, bumsub.ham\}@yonsei.ac.kr, dr.suhyun.kim@gmail.com
}

\usepackage{bibentry}

\begin{document}

\maketitle

\begin{abstract}
In vision transformers, position embedding (PE) plays a crucial role in capturing the order of tokens. However, in vision transformer structures, there is a limitation in the expressiveness of PE due to the structure where position embedding is simply added to the token embedding. A layer-wise method that delivers PE to each layer and applies independent Layer Normalizations for token embedding and PE has been adopted to overcome this limitation. In this paper, we identify the conflicting result that occurs in a layer-wise structure when using the global average pooling (GAP) method instead of the class token. To overcome this problem, we propose MPVG, which maximizes the effectiveness of PE in a layer-wise structure with GAP. Specifically, we identify that PE counterbalances token embedding values at each layer in a layer-wise structure. Furthermore, we recognize that the counterbalancing role of PE is insufficient in the layer-wise structure, and we address this by maximizing the effectiveness of PE through MPVG. Through experiments, we demonstrate that PE performs a counterbalancing role and that maintaining this counterbalancing directionality significantly impacts vision transformers. As a result, the experimental results show that MPVG outperforms existing methods across vision transformers on various tasks.
\end{abstract}

%

\section{Introduction}

Recently, vision transformers have become essential architecture in the field of computer vision due to their superior performance, surpassing CNNs in various tasks such as image classification, object detection, and semantic segmentation. This superiority has led to extensive research into numerous elements of vision transformer architecture, starting with ViT \cite{vit}.

Among the research on vision transformers, image representation methods for class prediction have been studied. In ViT, the class token is used to perform image representation, and the output of this token is then used to make class predictions via Multi-Layer Perceptron (MLP)~\cite {vit}. However, in several vision transformers, global average pooling (GAP) has been preferred over the class token method due to its translation-invariant characteristics and superior performance~\cite{cpvt}. As a result, the GAP method has been widely adopted in vision transformers~\cite{swin, twins, STViT, scaling}.

\begin{figure}[t]
\centering
\includegraphics[width=1.0\columnwidth]{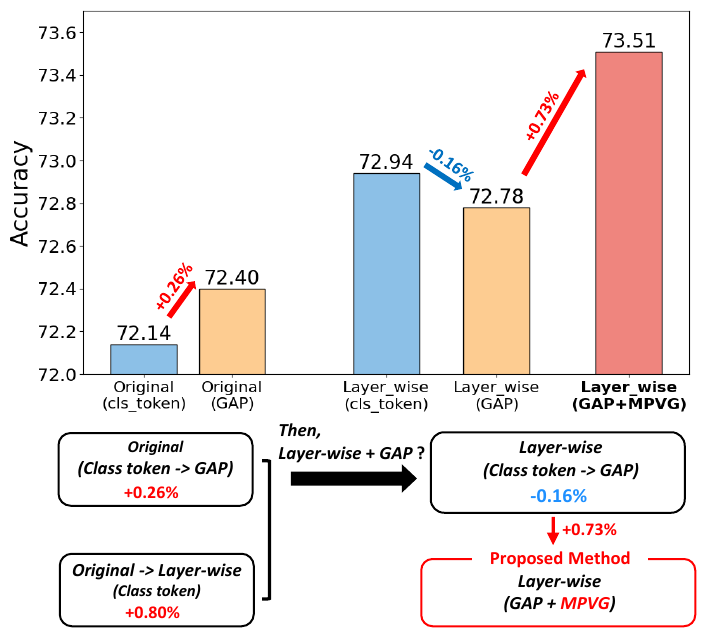}
\caption{
The conflicting result between the GAP method and the Layer-wise method. In DeiT-Ti, using the GAP method and the Layer-wise method separately results in performance improvements, but combining these two methods leads to a decrease in performance. As a result, MPVG resolves this phenomenon between the GAP and Layer-wise structure, maximizing the effect of PE.} \label{Figure1}

\end{figure}

Another research topic in vision transformers is position embedding (PE). PE plays a crucial role in providing positional information of tokens in the vision transformer, as the self-attention mechanism has an inherent deficiency in capturing the ordering of input tokens \cite{irpe, permutation}. In the original vision transformer, the expressiveness of the PE is limited due to its structure, where PE is simply added to the token embedding before being input into the first layer. To address this problem, each layer has independent Layer Normalizations (LNs) for the token embedding and PE, with PE being gradually delivered across all the layers~\cite{lape}. We refer to this method as "Layer-wise". The Layer-wise structure resolves the existing limitations and enhances the expressiveness of PE.

Interestingly, as shown in Fig~\ref{Figure1}, we observed results that differed from our expectations between the class token and GAP methods with PE delivered in the Layer-wise structure. On image classification, the GAP method demonstrates superior performance compared to the class token method~\cite{cpvt}. Additionally, the Layer-wise structure also improves the performance of vision transformers by enhancing the expressiveness of PE~\cite{lape}. However, we observed a conflicting result where performance decreased when the GAP and Layer-wise structure were applied together. Therefore, to overcome the conflicting results, we propose a method to maximize the effectiveness of PE in the GAP approach.

\begin{figure*}[ht!]
\centering
\includegraphics[width=2.0\columnwidth]{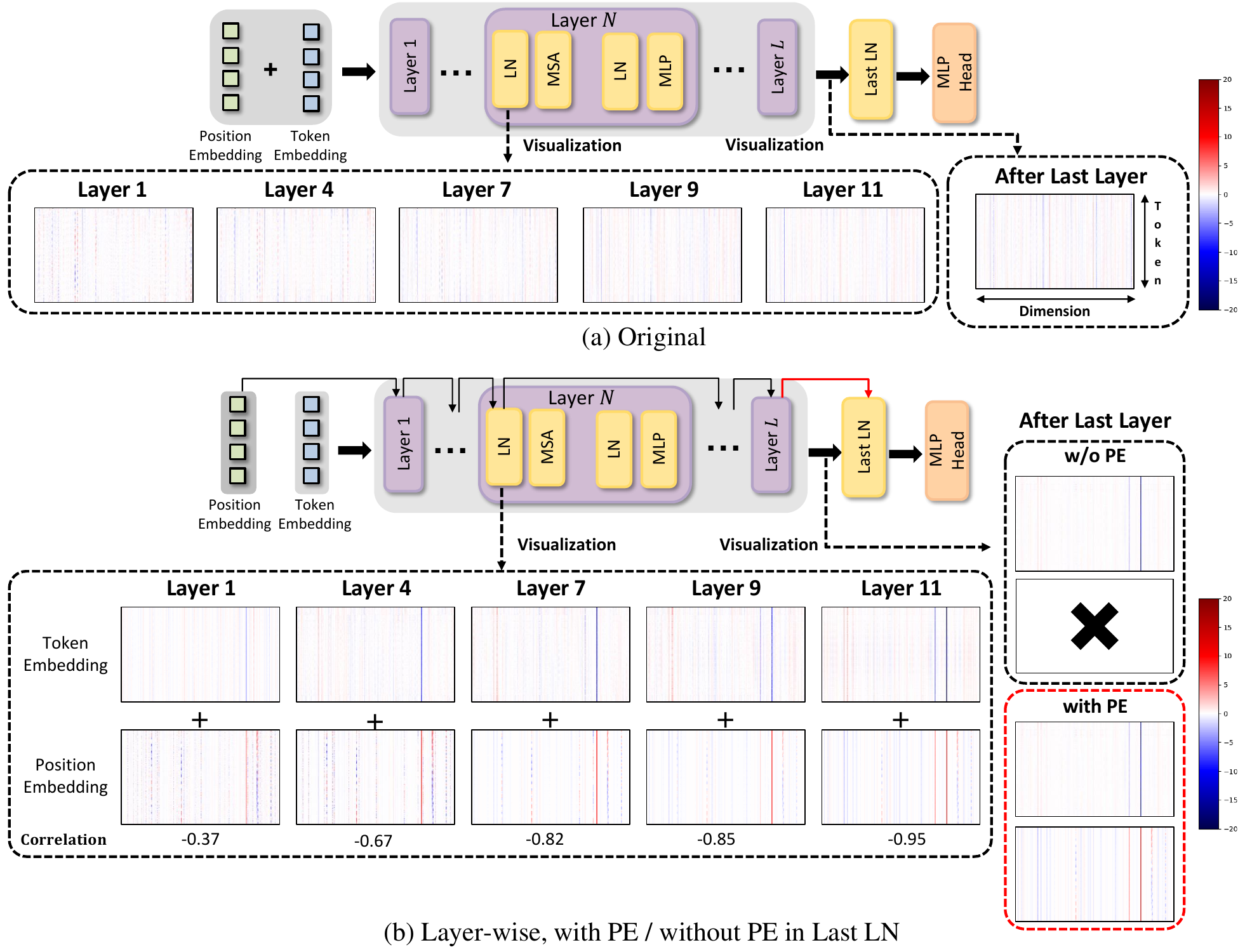}
\caption{The heatmaps depict the characteristics of each layer in both the original structure and the Layer-wise structure with the GAP method. For the Layer-wise structure, the heatmaps illustrate cases both with and without PE in the Last LN. For each heatmap based on DeiT-Ti, the x-axis represents the dimension of DeiT-Ti (256), and the y-axis represents the number of tokens (196). In both (a) and the top row (token embedding) of (b), the heatmaps represent the average value of token embedding in each layer, while the bottom row of (b) shows the heatmap of PE. The correlation in (b) refers to the correlation coefficient between token embedding and position embedding.}
\label{Figure2}
\end{figure*}

We observe that PE exhibits distinct characteristics at each layer in the Layer-wise structure, which are not seen in the original vision transformer. As shown in Fig~\ref{Figure2}, we find that PE tends to counterbalance the values of token embedding passing through the layers in the Layer-wise structure. Additionally, we observe that this tendency becomes more pronounced as the layers deepen. We also discover that in the Layer-wise structure, while the token embedding values maintain the counterbalanced effect by PE as they progress through the layers, as shown in Fig~\ref{Figure2}-(b), the directional balance of the token embedding is not adequately compensated by PE after passing through the last layer, even though it is still maintained. Through this observation, we establish two hypotheses: (1) in the Layer-wise structure, PE initially provides position information, but as the layers deepen, PE plays a role in counterbalancing the values of token embedding; (2) after the last layer, it is beneficial for vision transformers to maintain the directional balance by counterbalancing the token embedding values with PE.

To validate these hypotheses, we simply add PE to the Layer Normalization (LN) that exists outside the layers and before the MLP head. We call this LN as "Last LN". We refer to the method that uses an improved Layer-wise structure, different from the conventional Layer-wise structure, and does not deliver PE to the Last LN as PVG. Additionally, we refer to the method that maximizes the role of PE by delivering it to the Last LN as MPVG. By comparing PVG and MPVG, we demonstrate that MPVG effectively maximizes PE and that maintaining the counterbalancing directionality of PE is beneficial for vision transformers. Our experiments validate our hypothesis and demonstrate that MPVG outperforms other methods. The results demonstrate that MPVG consistently performs well for vision transformers.

In this paper, our contributions are as follows:

\begin{enumerate}
    \item We propose a simple yet effective method called MPVG, which maximizes the effect of PE in the GAP method. We show that MPVG leads to better performance in vision transformers.
    
    \item We provide an analysis of the phenomenon observed in PE when using the Layer-wise structure and offer insights into the role of PE.
    
    \item Through experiments, we verify that MPVG is generally effective for vision transformers on various tasks such as image classification, semantic segmentation, and object detection.
\end{enumerate}

\section{Related Work}

\subsection{Vision Transformers}
The vision transformer design is adapted from Transformer~\cite{trnasformer}, which was designed for natural language processing (NLP). This adaptation makes it suitable for computer vision tasks such as image classification~\cite{vit, DeiT, swin}, object detection~\cite{detr, detr2}, and semantic segmentation~\cite{setr, setr2, segmenter}. 

\subsubsection{Class Token \& Global Average Pooling}
ViT~\cite{vit} conducts ablation studies comparing the class token and GAP. Additionally, there are other studies on the use of GAP and class tokens in vision transformers~\cite{gap_cls, cpvt}. Studies such as CeiT~\cite{ceit} and T2T-ViT~\cite{t2t} use class token, while others like Swin Transformer~\cite{swin} and CPVT~\cite{cpvt} adapt GAP. CPVT achieves performance improvements by using GAP instead of the class token. Although the class token is not inherently translation-invariant, it can become so through training. By adopting GAP, which is inherently translation-invariant, better improvements in image classification tasks are achieved~\cite{cpvt}. Furthermore, GAP results in even less computational complexity because it eliminates the need to compute the attention interaction between the class token and the image patches. 

\subsection{Position Embeddings in Vision Transformers}

\subsubsection{Absolute Position Embedding} In the transformer, absolute position embedding is generated through a sinusoidal function and added to the input token embedding~\cite{trnasformer}. The sinusoidal functions are designed to give the position embedding locally consistent similarity, which helps vision transformers focus more effectively on tokens that are close to each other in the input sequence. This local consistency enhances the model's ability to capture spatial relationships and patterns~\cite{trnasformer}.

Besides sinusoidal positional embedding, position embedding can also be learnable. Learnable position embedding is created through training parameters, which are initialized with a fixed-dimensional tensor and updated along with the model's parameters during training. Recently, many models have adopted absolute position embedding due to their effectiveness in encoding positional information~\cite{vit, DeiT, swin}.

\subsubsection{Relative Position Embedding} In addition to absolute position embedding, there is also relative position embedding~\cite{rpe}. Relative PE encodes the relative position information between tokens. The first to propose relative PE in computer vision was \cite{rpe}. Furthermore, \cite{2drpe} proposed a 2-D relative position encoding for image classification that showed superior performance compared to traditional 2-D sinusoidal embedding. This relative encoding captures spatial relationships between tokens more effectively. In related research, iRPE \cite{irpe} improves relative PE by incorporating query interactions and relative distance modeling in self-attention. RoPE \cite{rope} introduces flexible sequence lengths, decaying inter-token dependency, and relative position encoding in linear self-attention.

\begin{figure*}[t]
\centering
\includegraphics[width=2.0\columnwidth]{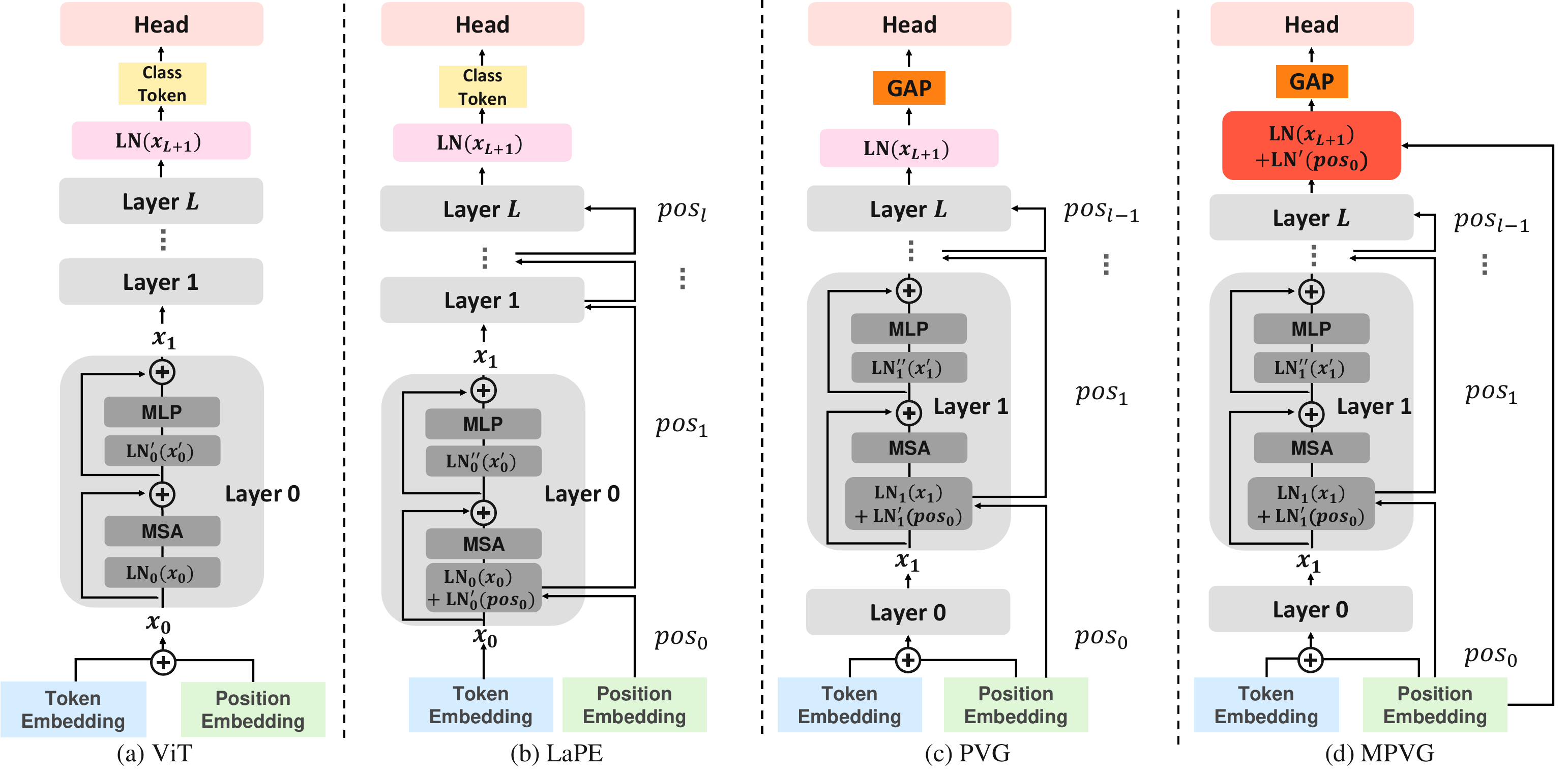}
\caption{The overview of the various methods. (a) ViT. (b) LaPE~\cite{lape}. (c) PVG, an improved Layer-wise structure. Specifically, we adopt a structure where the token embedding and PE are added before entering layer 0 and a hierarchical structure for delivering PE, excluding layer 0. (d) MPVG. The main difference from PVG is whether the initial PE is delivered to the Last LN.} \label{Figure3}

\end{figure*}

\section{Methodology}
In this section, we first explain the absolute position embedding and then provide a detailed overview of the Layer-wise structure~\cite{lape}. Next, we introduce PVG, an improved Layer-wise structure, along with MPVG, which effectively leverages the characteristics of PE in the Layer-wise structure.

\subsection{Preliminary: Absolute Position Embedding}
The method of absolute position embedding used in vision transformers is as follows. As shown in Fig~\ref{Figure3}-(a), PE is added to the token embedding before they are input into the layer. This can be expressed as follows:
\begin{equation}
x_0 = [x_{\text{cls}}; \; p^1; \;p^2; \; \ldots \; p^N;] + pos, \tag{1}
\label{equation1}
\end{equation} 

\noindent where $\mathit{p}$ and $\mathit{pos}$ represent the patch and position embedding, respectively. \(N\) represents the number of patches, calculated as \(HW / P^2\), where \(H\) and \(W\) are the height and width of the image, and \(P \times P\) is the resolution of each patch. The combined token embedding and PE, denoted as $\mathit{x}$, can be expressed in a layer as follows:

\begin{equation}
x'_l = \text{MSA}(\text{LN}_{l}(x_{l})) + x_{l} \quad (l = 0 \ldots L), \tag{2}
\label{equation2}
\end{equation}

\begin{equation}
x_{l+1} = \text{MLP}(\text{LN}^{'}_{l}(x'_l)) + x'_l \quad (l = 0 \ldots L), \tag{3}
\label{equation3}
\end{equation}

\begin{equation}
y = \text{LN}(x_{L+1}) \tag{4}
\label{equation4}
\end{equation}

\noindent where LN, LN', and LN'' represent different Layer Normalizations, Multi-head Self-Attention is denoted as MSA, and Multi-Layer Perceptron is denoted as MLP. $x_{L+1}$ refers to the value after passing through the last layer $L$.

\subsection{Preliminary: Layer-wise Structure}

LaPE~\cite{lape} points out problems with the joining method that position embedding and token embedding in the vision transformers. As shown in Eq.~(\ref{equation1}), when PE is added to the token embedding before the first layer, and the same LN is applied to both the token embedding and PE as in Eq.~(\ref{equation2}), they share the same affine parameters in LN. This method limits the expressiveness of PE. Therefore, the Layer-wise structure is used to resolve these problems. This can be expressed as follows:

\begin{equation}
x_0 = [x_{\text{cls}}; \; p^1; \;p^2; \; \ldots \; p^N;], \tag{5}
\label{equation5}
\end{equation} 

\begin{equation}
x'_l = \text{MSA}(\text{LN}_{l}(x_{l}) + \text{LN}^{'}_{l}(pos_{l})) + x_{l} \tag{6}
\label{equation6}
\end{equation}

\noindent Eq.~(\ref{equation1}) is modified to Eq.~(\ref{equation5}) and Eq.~(\ref{equation2}) to Eq.~(\ref{equation6}). In Eq.~(\ref{equation6}), the Layer-wise structure uses independent LN for token embedding($\mathit{x}$) and PE. PE is delivered in each layer as follows :

\begin{equation} 
\left\{
\begin{aligned}
    pos_{0} &= pos \\
    pos_{l} &= \text{LN}^{'}_{l-1}(pos_{l-1}) \quad (l = 1 \ldots L)
\end{aligned}
\right.
\tag{7}
\label{equation7}
\end{equation}

\subsection{Maximizing the Position Embedding with GAP}

In this section, we propose two methods, MPVG and PVG, to validate our hypothesis. In Fig \ref{Figure2}-(b), we observed that, in Layer-wise structure, the effect of PE in counterbalancing the values of token embedding($\mathit{x}$) becomes more pronounced as the layers deepen, as evidenced by the correlation between the two. However, in Layer-wise structure, although the directionality of the token embedding is maintained outside the layer, there is no PE to counterbalance that value. Therefore, we validate our hypothesis by comparing MPVG, which delivers PE to the Last LN, with PVG, which does not.

We remove the class token as we adapt the Global Average Pooling (GAP) method. Although we use the Layer-wise structure, we modify specific details. Specifically, we combine two structural approaches: (1) adding token embedding and PE before inputting the layer. (2) delivering PE to each layer except the 0th layer. We call this method as PVG. In PVG method, as shown in Figure~\ref{Figure3}-(c), is as follows:

\begin{equation}
x_0 = [p^1; \;p^2; \; \ldots \; p^N;] + pos, \tag{8}
\label{equation8}
\end{equation} 

\begin{equation}
x_l' = 
\begin{cases} 
\text{MSA}(\text{LN}_{0}(x_0)) + x_0 & \text{if } l = 0 \\ 
\text{MSA}(\text{LN}_{l}(x_{l}) + \text{LN}'_{l}(pos_{l-1})) + x_{l} & \text{if } 1 \leq l \leq L \tag{9}
\end{cases}
\label{equation9}
\end{equation}

\begin{equation} 
\left\{
\begin{aligned}
    pos_{0} &= pos \\
    pos_{l} &= \text{LN}^{'}_{l}(pos_{l-1}) \quad (l = 1 \ldots L)
\end{aligned}
\right.
\tag{10}
\label{equation10}
\end{equation}

\noindent The subsequent process is the same as in Eq.~(\ref{equation3}) and Eq.~(\ref{equation4}). In MPVG, we modify Eq.~(\ref{equation4}) as follows after going through the process of PVG:

\begin{equation}
y = \text{LN}(x_{L+1}) + \text{LN}^{'}(pos_0) \tag{11}
\label{equation11}
\end{equation}

To verify whether maintaining the counterbalance effect of PE is beneficial, we deliver PE to the Last LN in PVG, as shown in Eq.~(\ref{equation11}). We refer to this method as MPVG. In the next section, we verify the superiority of MPVG by comparing the two methods. Also, we show that MPVG outperforms previous approaches through experiments across various vision transformers and datasets.

\section{Experiment}

\subsubsection{Training Settings} All experiments are conducted on an RTX 4090 with 4 GPUs using AdamW optimizer~\cite{adamw}, while DeiT-B is trained on an RTX 4090 with 8 GPUs. 

\subsection{Image Classification}

We evaluate the performance of our methods on ImageNet-1K~\cite{imagenet} and CIFAR-100~\cite{cifar100}. On ImageNet-1K, we conduct experiments with DeiT~\cite{DeiT}, Swin~\cite{swin}, CeiT~\cite{ceit}, and T2T-ViT~\cite{t2t}. In the case of Swin, due to its staged architecture that generates hierarchical representations with the same feature map resolution as convolutional networks, both PVG and MPVG exceptionally include layer 0. All vision transformers are trained on 224×224 resolution images for 300 epochs, except T2T-ViT-7, which is trained for 310 epochs. 

On CIFAR-100, we conduct experiments using ViT-Lite~\cite{cct} and T2T-ViT-7~\cite{t2t}. ViT-Lite was trained for 310 epochs on 32×32 resolution images with a batch size of 128. In the case of T2T-ViT-7, we transfer our pretrained T2T-ViT to downstream datasets such as CIFAR-100 and finetune the pretrained T2T-ViT-7 for 60 epochs with a batch size of 128.
\begin{table}[t]
\renewcommand\arraystretch{1.2}
\resizebox{\columnwidth}{!}{

\begin{tabular}{cccc}
\hline 
Model                                                                                                             & Method        & \#Params (M) & \begin{tabular}[c]{@{}c@{}}Top-1 \\ Acc (\%)\end{tabular} \\ \hline \hline
\multicolumn{1}{c|}{\multirow{4}{*}{\begin{tabular}[c]{@{}c@{}}DeiT-Ti\\ \cite{DeiT}\end{tabular}}}  & Default       & 5.717        & 72.14                                                     \\
\multicolumn{1}{c|}{}                                                                                             & LaPE          & 5.721        & 72.94                                                     \\
\multicolumn{1}{c|}{}                                                                                             & PVG           & 5.721        & 73.17                                                     \\
\multicolumn{1}{c|}{}                                                                                             & \textbf{MPVG} & 5.721        & \textbf{73.51}                                            \\ \hline
\multicolumn{1}{c|}{\multirow{4}{*}{\begin{tabular}[c]{@{}c@{}}DeiT-S\\ \cite{DeiT}\end{tabular}}}   & Default       & 22.050       & 79.81                                                     \\
\multicolumn{1}{c|}{}                                                                                             & LaPE          & 22.059       & 80.39                                                     \\
\multicolumn{1}{c|}{}                                                                                             & PVG           & 22.058       & 80.38                                                     \\
\multicolumn{1}{c|}{}                                                                                             & \textbf{MPVG} & 22.059       & \textbf{80.61}                                            \\ \hline
\multicolumn{1}{c|}{\multirow{4}{*}{\begin{tabular}[c]{@{}c@{}}DeiT-B\\ \cite{DeiT}\end{tabular}}}   & Default       & 86.567       & 81.85                                                     \\
\multicolumn{1}{c|}{}                                                                                             & LaPE          & 86.586       &  82.15                                                         \\
\multicolumn{1}{c|}{}                                                                                             & PVG           & 86.583             &  82.21                                                         \\
\multicolumn{1}{c|}{}                                                                                             & \textbf{MPVG} &  86.584            &   \textbf{82.42}                                                        \\ \hline
\multicolumn{1}{c|}{\multirow{4}{*}{\begin{tabular}[c]{@{}c@{}}Swin-Ti\\ \cite{swin}\end{tabular}}}  & Default       & 28.589       & 81.37                                                     \\
\multicolumn{1}{c|}{}                                                                                             & LaPE          & 28.599       & 81.48                                                     \\
\multicolumn{1}{c|}{}                                                                                             & PVG           & 28.598       & 81.52                                                     \\
\multicolumn{1}{c|}{}                                                                                             & \textbf{MPVG} & 28.599       & \textbf{81.64}                                            \\ \hline
\multicolumn{1}{c|}{\multirow{4}{*}{\begin{tabular}[c]{@{}c@{}}CeiT-Ti\\ \cite{ceit}\end{tabular}}}  & Default       & 6.356        & 76.62                                                     \\
\multicolumn{1}{c|}{}                                                                                             & LaPE          & 6.361        & 76.89                                                     \\
\multicolumn{1}{c|}{}                                                                                             & PVG           & 6.361        & 77.14                                                     \\
\multicolumn{1}{c|}{}                                                                                             & \textbf{MPVG} & 6.361        & \textbf{77.20}                                            \\ \hline
\multicolumn{1}{c|}{\multirow{4}{*}{\begin{tabular}[c]{@{}c@{}}T2T-ViT-7\\ \cite{t2t}\end{tabular}}} & Default       & 4.310        & 71.76                                                     \\
\multicolumn{1}{c|}{}                                                                                             & LaPE          & 4.313        & 72.01                                                     \\
\multicolumn{1}{c|}{}                                                                                             & PVG           & 4.312        & 71.91                                                     \\
\multicolumn{1}{c|}{}                                                                                             & \textbf{MPVG} & 4.313        & \textbf{72.28}                                            \\ \hline
\end{tabular}
}
\caption{Top-1 accuracy comparison with various methods, using DeiT-T, DeiT-S, DeiT-B, Swin-Ti, CeiT-Ti, T2T-ViT-7 on ImageNet-1K.}
\label{table1}
\end{table}

\begin{table}[t!]
\resizebox{\columnwidth}{!}{
\begin{tabular}{cccc}
\hline 
Model                                                                                                             & \begin{tabular}[c]{@{}c@{}}Method\end{tabular} & \#Param (M) & \begin{tabular}[c]{@{}c@{}}Top-1\\ Acc (\%)\end{tabular} \\ \hline \hline
\multicolumn{1}{c|}{\multirow{4}{*}{\begin{tabular}[c]{@{}c@{}}ViT-Lite\\ \cite{cct}\end{tabular}}} & Default                                             & 3.740       & 74.90                                                    \\
\multicolumn{1}{c|}{}                                                                                             & LaPE                                                & 3.744       & 75.52                                                    \\
\multicolumn{1}{c|}{}                                                                                             & PVG                                                 & 3.742       & 76.67                                                    \\
\multicolumn{1}{c|}{}                                                                                             & \textbf{MPVG}                                       & 3.743       & \textbf{76.87}                                           \\ \hline
\multicolumn{1}{c|}{\multirow{4}{*}{\begin{tabular}[c]{@{}c@{}}T2T-ViT-7\\ \cite{t2t}\end{tabular}}} & Default                                             &  4.078    &   83.22                                              \\
\multicolumn{1}{c|}{}                                                                                             & LaPE                                                &  4.082     &    83.41                                               \\
\multicolumn{1}{c|}{}                                                                                             & PVG                                                 &  4.081    &   83.39                                                  \\
\multicolumn{1}{c|}{}                                                                                             & \textbf{MPVG}                                       &  4.081     & \textbf{83.51}                                           \\ \hline
\end{tabular}
}
\caption{Top-1 accuracy comparison with various methods, using ViT-Lite and T2T-ViT-7 on CIFAR-100. In the case of T2T-ViT, the results are based on fine-tuning the pretrained model on the downstream dataset, CIFAR-100.}
\label{table2}
\end{table}

As shown in Table~\ref{table1}, For MPVG, the performance on DeiT-Ti improved from 72.14\% to 73.51\%, representing an increase of approximately 1.37\%. For DeiT-S, the performance improved from 79.81\% to 80.61\%, an increase of approximately 0.80\%. Additionally, there were performance improvements of 0.57\% in DeiT-B, 0.27\% in Swin-Ti, 0.58\% in CeiT, and 0.52\% in T2T-ViT. Overall, MPVG outperforms the existing methods in all cases. Moreover, we confirm that MPVG consistently demonstrates superior performance compared to PVG across various vision transformers.

As shown in Table~\ref{table2}, MVPG achieves overall performance improvements on CIFAR-100. Specifically, MPVG improves the performance of ViT-Lite by 1.97\%, from 74.90\% to 76.87\%, and enhances the performance of T2T-ViT-7 by 0.29\% over the default. Additionally, MPVG shows a 0.2\% and 0.12\% improvement over PVG for ViT-Lite and T2T-ViT-7, respectively. Overall, MPVG outperforms existing methods across all cases on CIFAR-100.

\subsection{Object Detection}
On object detection, we evaluate our methods on COCO 2017~\cite{coco}. To demonstrate the effectiveness of our method on object detection tasks, we select the ViT-Adapter-Ti~\cite{vitadapter} model based on Mask R-CNN~\cite{mask-rcnn} in MMDetection framework~\cite{mmdet}. Additionally, we use the default settings and train it for 36 epochs using the 3x+MS(multi-scale training) schedule. As shown in Table~\ref{tabel3}, MPVG achieves improvements of +0.6 in box AP and +0.5 in mask AP compared to the default setting. MPVG, in particular, demonstrates superior performance with an increase of +0.5 in box AP and +0.4 in mask AP over PVG.

\subsection{Semantic Segmentation}
On semantic segmentation, we evaluate our methods on ADE20K~\cite{ade20k}. We select the ViT-Adapter-Ti~\cite{vitadapter} model based on UperNet~\cite{upernet} in MMsegmentation framework~\cite{mmseg} and train it using the default settings. As shown in Table~\ref{table4}, MPVG achieves an improvement of +1.14 mIoU compared to the default. Furthermore, MPVG outperforms PVG, achieving a performance improvement of +0.62 mIoU.

\begin{table}[t]
\resizebox{\columnwidth}{!}{
\begin{tabular}{cccc}
\hline
Model                                                & Pre-trained                                      & Method & AP$^{\text{box}}$ / AP$^{\text{mask}}$ \\ \hline\hline
\multicolumn{1}{c|}{\multirow{4}{*}{ViT-Adapter-Ti}} & \multicolumn{1}{c|}{\multirow{4}{*}{DeiT-Ti}} & \multicolumn{1}{c}{Default}   & 45.9 / 41.0                       \\
\multicolumn{1}{c|}{}                                & \multicolumn{1}{c|}{}                         & \multicolumn{1}{c}{LaPE}      & 46.2 / 41.2                                \\ 
\multicolumn{1}{c|}{}                                & \multicolumn{1}{c|}{}                         & \multicolumn{1}{c}{PVG}       & 46.1 / 41.2                                  \\
\multicolumn{1}{c|}{}                                & \multicolumn{1}{c|}{}                         & \multicolumn{1}{c}{\textbf{MPVG}}      &  \textbf{46.5} / \textbf{41.4}                                 \\ \hline
\end{tabular}
}
\caption{Performance comparison of Object Detection on COCO2017. For comparison, DeiT-Ti model pretrained on ImageNet-1K with each method is used.}
\label{tabel3}
\end{table}

\begin{table}[t]
\centering{
\begin{tabular}{cccc}
\hline
Model                                                & Pre-trained                                      & Method & mIoU \\ \hline\hline
\multicolumn{1}{c|}{\multirow{4}{*}{ViT-Adapter-Ti}} & \multicolumn{1}{c|}{\multirow{4}{*}{DeiT-Ti}} & \multicolumn{1}{c}{Default}   & 40.55 \\
\multicolumn{1}{c|}{}                                & \multicolumn{1}{c|}{}                         & \multicolumn{1}{c}{LaPE}      & 41.42 \\ 
\multicolumn{1}{c|}{}                                & \multicolumn{1}{c|}{}                         & \multicolumn{1}{c}{PVG}       & 41.07 \\
\multicolumn{1}{c|}{}                                & \multicolumn{1}{c|}{}                         & \multicolumn{1}{c}{\textbf{MPVG}}      & \textbf{41.69} \\ \hline
\end{tabular}
}
\caption{Performance comparison of Semantic Segmentation on ADE20K. For comparison, DeiT-Ti model pretrained on ImageNet-1K with each method is used.}
\label{table4}
\end{table}

\begin{figure}[t!]
\centering
\includegraphics[width=1.0\columnwidth]{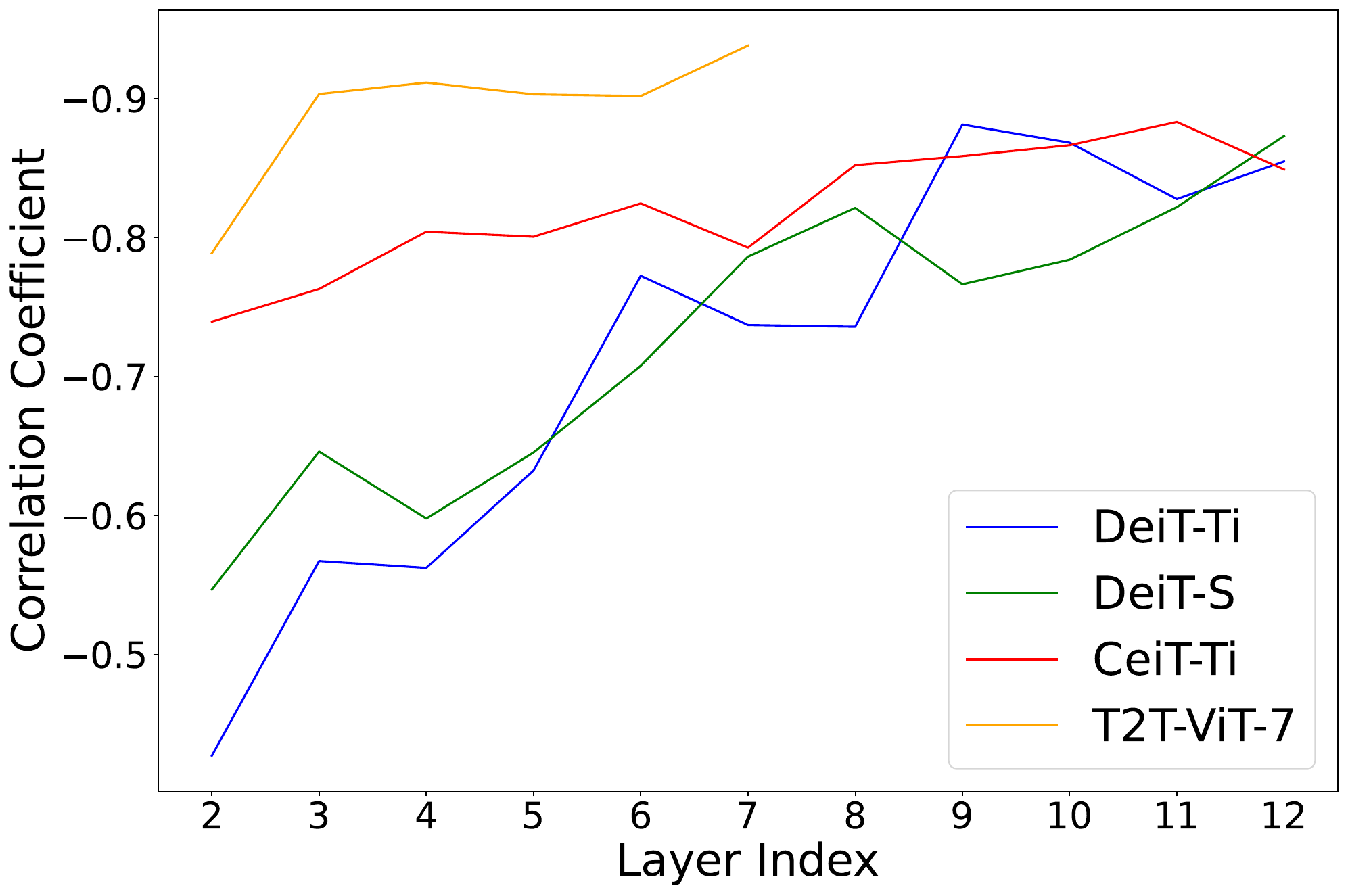}
\caption{Correlation coefficient between token embedding and position embedding in Layer-wise. Each token embedding and position embedding is based on the values after applying LN. DeiT-Ti, DeiT-S, and CeiT-Ti each have a total of 12 layers, but T2T-ViT-7 has 7 layers.
 } \label{Figure4}

\end{figure}

\begin{figure}[t]
\centering
\includegraphics[width=1.0\columnwidth]{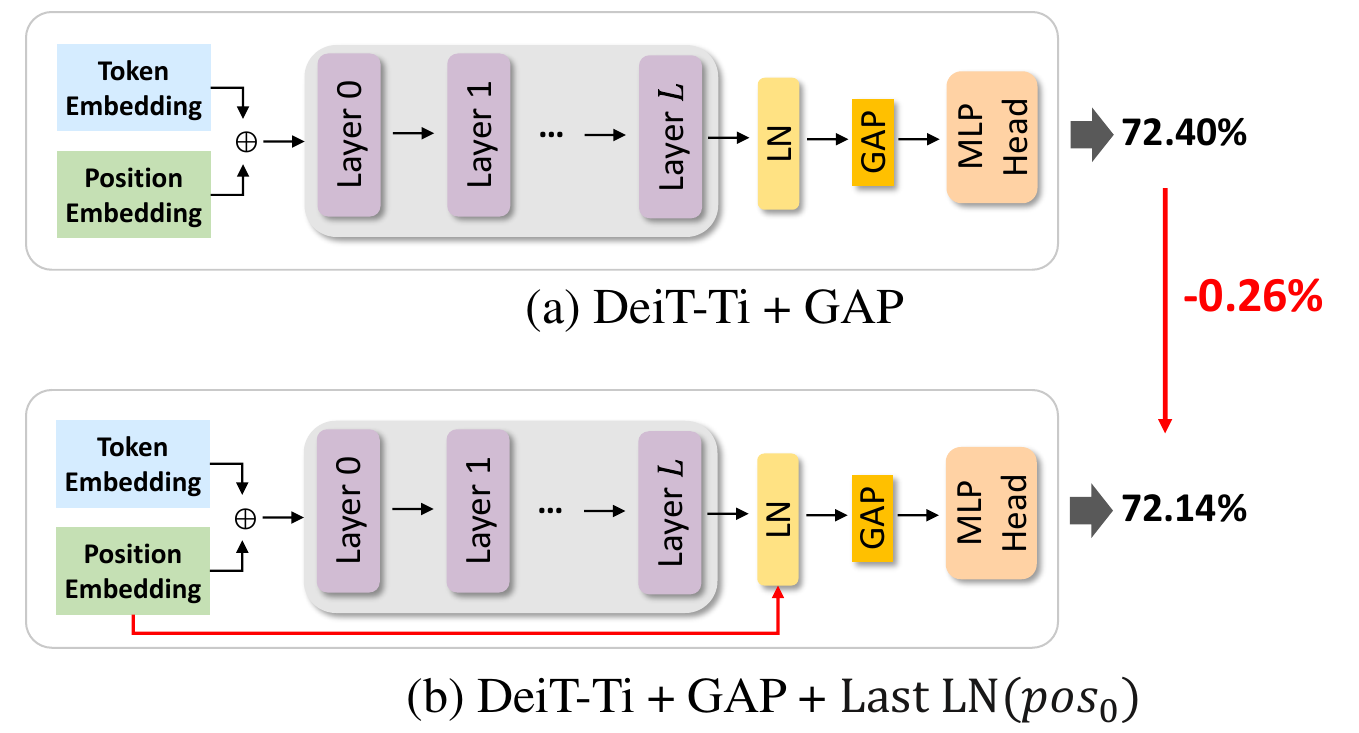}
\caption{Comparison of two methods on DeiT-Ti. (a) Structure with only GAP applied, showing 72.40\% performance; and (b) Structure with GAP and position embedding added to the Last LN in a non-Layer-wise structure, also showing 72.14\% performance. 
 } \label{Figure5}
\end{figure}

\subsection{Analysis}

Through experiments on image classification, object detection, and semantic segmentation, we demonstrate the effectiveness of MPVG. In all tasks, MPVG not only outperforms the baseline but also achieves the best performance among all methods. This validates our hypothesis and proves that our method is an effective approach to maximizing PE in the GAP method. Fig~\ref{Figure4} shows that in Layer-wise structure, token embedding and position embedding exhibit increasingly opposing directions as the layers deepen. This suggests that PE not only provides positional information in the initial layers but also may play a counterbalancing role that becomes more pronounced in deeper layers. To further explore this, we compare PVG and MPVG to confirm that PE has a counterbalancing effect. This comparison proves that maintaining the counterbalancing role of PE impacts the performance of vision transformers.

In conclusion, several key points can be identified: (1) In the initial layers, PE primarily provides positional information, enabling the model to understand the spatial relationships between tokens. However, as the layers deepen, PE plays a role in counterbalancing the token embedding. (2) This counterbalancing effect of PE has a significant impact on the performance of vision transformers. Therefore, MPVG demonstrates that maintaining this direction is beneficial for vision transformers and proves that PE can perform additional roles to sustain this effect.

\begin{table}[t!]
\resizebox{\columnwidth}{!}{
\begin{tabular}{cccc}
\hline 
Model                                         & PE Method                                  & Last LN                              & Top-1 Accuracy (\%) \\ \hline\hline
\multicolumn{1}{c|}{\multirow{4}{*}{DeiT-Ti}} & \multicolumn{1}{c|}{\multirow{4}{*}{MPVG}} & \multicolumn{1}{c|}{$pos_{11}$} & 73.30      \\
\multicolumn{1}{c|}{}                         & \multicolumn{1}{c|}{}                      & \multicolumn{1}{c|}{$pos_{8}$}          &       73.38              \\
\multicolumn{1}{c|}{}                         & \multicolumn{1}{c|}{}                      & \multicolumn{1}{c|}{$pos_{5}$}          &     73.39                \\
\multicolumn{1}{c|}{}                         & \multicolumn{1}{c|}{}                      & \multicolumn{1}{c|}{\bm{$pos_{0}$}}         & \textbf{73.51}              \\ \hline
\end{tabular}
}
\caption{Comparison of the value of PE added to the Last LN in MPVG. $pos_{0}$ refers to the initial position embedding, and $pos_{11}$ represents the position embedding after applying LN in the last layer. $pos_{N}$ (=$\text{LN}^{'}_{N}(pos_{N-1})$) indicates the PE input for the $\mathit{(N+1)}$th layer.}
\label{table5}
\end{table}

\subsection{Effect of PE in Last LN}
We conduct additional experiments to validate our hypothesis. Specifically, we aim to confirm that adding PE to the Last LN effectively maintains the counterbalancing role of PE in Layer-wise structure. We compare the method using only GAP with the method that adds PE to the Last LN in a non-Layer-wise structure while using GAP. We perform these experiments with DeiT-Ti on ImageNet-1K.

In Fig 5, we compare (a), where only GAP is applied, with (b), where PE is delivered to the Last LN in a non-Layer-wise structure with GAP. Fig~\ref{Figure5}-(a) shows a 72.40\% performance, while Fig~\ref{Figure5}-(b) shows a decreased performance of 72.14\%. This indicates that adding PE to the Last LN is only effective in Layer-wise structure where PE is delivered to each layer. In other words, these experimental results prove that in Layer-wise structure, the token embedding contains values that are counterbalanced by PE. Specifically, in Fig 5-(b) structure, the token embedding that passes through the layers does not contain the directional values, which is counterbalanced by PE. Thus, adding PE to the Last LN not only has no effect but actually leads to a decrease in performance. As a result, as shown in Fig 2-(b), in Layer-wise structure, the token embedding progresses while retaining values that are meant to be counterbalanced by PE, but after passing through the final layer, this directional value is not adequately compensated by PE. This proves that PE is necessary to perform this additional counterbalancing role in the Last LN.

\subsection{Ablation Study}

\subsubsection{The impact of PE values delivered to the Last LN} We conduct experiments to investigate the impact of varying the PE values passed to the Last LN in MPVG. In Table~\ref{table5}, \(\mathit{pos}_{N}\) represents the value of \(\text{LN}^{'}_{N}(\mathit{pos}_{N-1})\). Since MPVG does not deliver PE to layer 0, \(N\) ranges from 1 to \(L-1\), where \(L\) is the number of layers. Experiments show that MPVG consistently outperforms PVG, which achieves a performance of 73.17\%, regardless of the PE values passed to the Last LN. This suggests that delivering PE in the Last LN has a significant positive impact on the performance of vision transformers. Furthermore, it demonstrates that maintaining the role of PE in the Last LN is generally effective. MPVG adopts \(\mathit{pos}_{0}\), which shows the best performance by comparing various PE values delivered to the Last LN.

\begin{table}[t]
\renewcommand\arraystretch{1.0}
\resizebox{\columnwidth}{!}{
\begin{tabular}{ccccc}
\hline 
\multirow{2}{*}{Model}                     & \multicolumn{3}{c}{Structure} & \multirow{2}{*}{\begin{tabular}[c]{@{}c@{}}Top-1\\ Acc (\%)\end{tabular}} \\
                                           & Layer 0  & Hierarchical  & ($\mathit{x}$+PE) &                                                                           \\ \hline\hline
\multicolumn{1}{c|}{\multirow{4}{*}{MPVG}} & \ding{55}       & \checkmark             & \checkmark       &    73.31                                                                       \\
\multicolumn{1}{c|}{}                      & \checkmark       & \ding{55}             & \checkmark       &       73.48                                                                    \\
\multicolumn{1}{c|}{}                      & \checkmark       & \checkmark             & \ding{55}       &       73.28                                                                    \\
\multicolumn{1}{c|}{}                      & \checkmark       & \checkmark             & \checkmark       & \textbf{73.51}                                                            \\ \hline
\end{tabular}
}
\caption{Structural Differences in MPVG. "Layer 0" denotes whether layer 0 is included when delivering PE to layers. ``Hierarchical'' denotes whether \(\mathit{pos}_l\) is \(\text{LN}^{'}_{l}(\mathit{pos}_{l-1})\) or \(\text{LN}^{'}_{l}(\mathit{pos}_{0})\).
 "($\mathit{x}$+PE)" denotes whether PE is added to the token embedding($\mathit{x}$) before entering layer 0 or not.}
\label{table6}
\end{table}

\subsubsection{Structural differences in MPVG} We also experiment by varying the architecture structure in MPVG. Table 6 presents the ablations for the differences in architecture within MPVG. The experiments are conducted on DeiT-Ti using the ImageNet-1K. Through this experiment, we adopt an improved Layer-wise structure that differs from the conventional Layer-wise structure.

Specifically, we conduct comparative experiments on three structural differences: (1) Our methods exclude layer 0 when delivering PE. Through our experiments, we find that delivering PE to layer 0, which was previously included, is not only unnecessary but also improves the performance of vision transformers when excluded. (2) We add PE to the token embedding before it enters layer 0. Unlike LaPE where token embedding $\mathit{x}$ and PE are separated before entering the first layer, our methods add PE to $\mathit{x}$ before entering layer 0. This structure does not limit the expressiveness of PE because independent LN is applied to both the token embedding and PE in each layer, and PE is delivered in a Layer-wise structure. Moreover, the ($\mathit{x}$+PE) structure boosts performance by approximately 0.23\%. (3) We observe that the performance is similar between hierarchical and non-hierarchical structures. However, in non-hierarchical structures, performance often declines in small or large vision transformers due to overfitting~\cite{lape}. Through Table 6, we demonstrate that our methods represent the optimal structure in the Layer-wise structure.

\section{Conclusion}
We reveal that position embedding can play additional roles in vision transformers using the GAP method. Specifically, in a Layer-wise structure, PE has a counterbalancing effect on the values of token embedding, and maintaining this directional balance by PE is beneficial for vision transformers. Based on these observations, we propose a simple yet effective method, MPVG. MPVG utilizes the characteristics of PE observed in the Layer-wise structure to maximize the PE. Through extensive experiments, we demonstrate that MPVG is generally effective on vision transformers, outperforming previous methods. However, MPVG has a potential limitation in that it is incompatible with the class token method. Through these limitations, we will further explore the broader applicability of MPVG and the effects of PE's counterbalancing as part of our future work. In this paper, we demonstrate that MPVG effectively addresses the issues arising in GAP and Layer-wise structures, providing a significantly more meaningful approach. Through this, we look forward to MPVG offering a broader perspective on position embedding.

\section{Acknowledgments}
This research was supported by the MSIT(Ministry of Science and ICT), Korea, under the ITRC(Information Technology Research Center) support program(IITP-2024-RS-2023-00258649, 80\%) supervised by the IITP(Institute for Information \& Communications Technology Planning \& Evaluation) and was partly supported by the IITP grant funded by the Korea government (MSIT) (No.RS-2022-00143524, Development of Fundamental Technology and Integrated Solution for Next-Generation Automatic Artificial Intelligence System) and (No.RS2023-00225630, Development of Artificial Intelligence for Text-based 3D Movie Generation).

\bibliography{aaai25}

%

%

\lstset{%
	basicstyle={\footnotesize\ttfamily},
	numbers=left,numberstyle=\footnotesize,xleftmargin=2em,
	aboveskip=0pt,belowskip=0pt,%
	showstringspaces=false,tabsize=2,breaklines=true}
\floatstyle{ruled}
\newfloat{listing}{tb}{lst}{}
\floatname{listing}{Listing}
%
\pdfinfo{
/TemplateVersion (2025.1)
}

\setcounter{secnumdepth}{0} 

\twocolumn[{%
    \renewcommand\twocolumn[1][]{#1}%
    \vspace{-3em}
    \begin{center}
        \centering
        \includegraphics[width=0.99\textwidth]{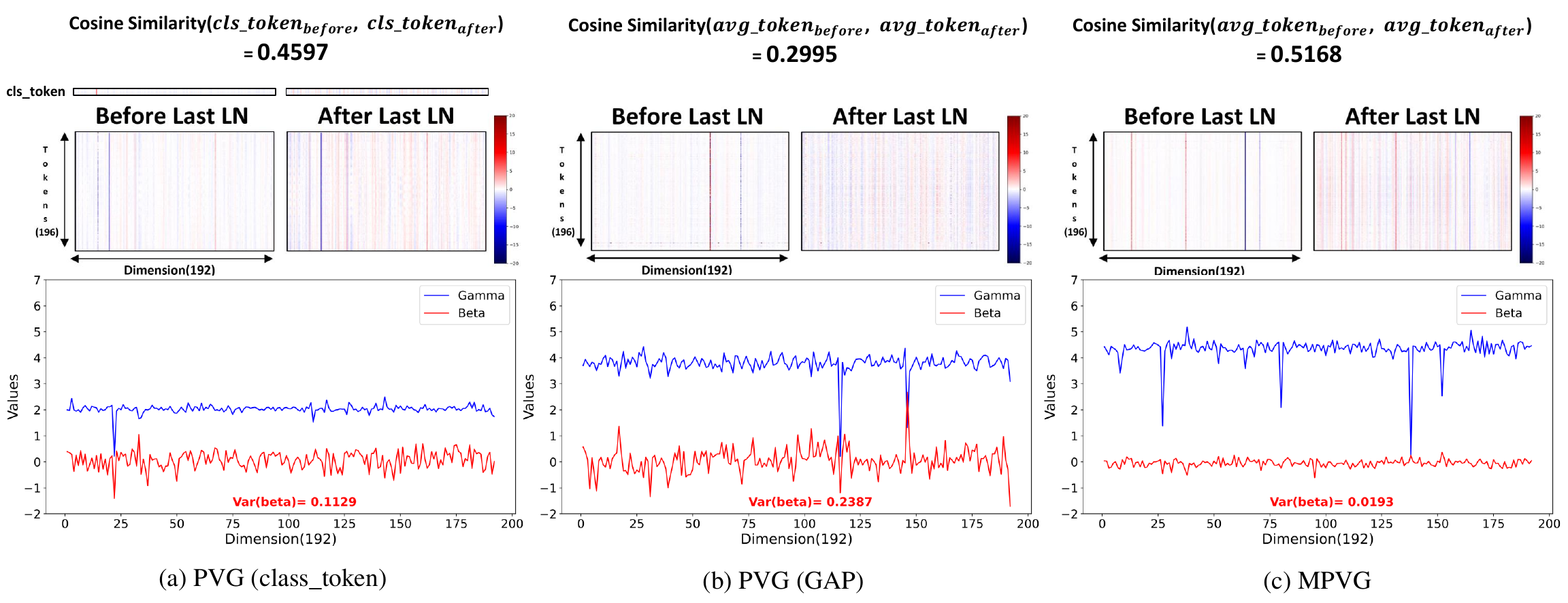}
        \captionof{figure}{
        Heatmaps(averaged over the batch size), beta and gamma values in the Last LN, and cosine similarity for each method. (a) represents the PVG using the class token. The cosine similarity refers to the similarity of the class token before and after the Last LN. (b) represents the PVG using GAP. The cosine similarity refers to the similarity of the token averages in the heatmaps before and after the Last LN. (c) represents the MPVG. The cosine similarity refers to the similarity of the token averages in the heatmaps before and after the Last LN.
        }
        \label{supple_fig1}
        \vspace{5pt}
    \end{center}%
    }]
    
\section{Appendix}

\subsection{A Detailed Analysis of PE in the Last LN}
We provide a detailed analysis of the role of position embedding (PE) in the Last LN (LN means Layer Normalization~\cite{ln}). As shown in Fig~\ref{supple_fig1}, (b) visualizes the gamma and beta values, which are the affine parameters of the Last LN, in PVG where PE is not delivered to the Last LN. In (c), the gamma and beta values of the Last LN are visualized in MPVG, where PE is delivered to the Last LN.

In PVG, upon examining the beta affine parameter, we find that the variance of beta is 0.2387, indicating significant variability. Specifically, we observe that the beta value counterbalances the high-value dimensions present before the Last LN. In contrast, in MPVG, where PE is delivered to the Last LN, the variance of beta is much smaller at 0.0193 compared to PVG. This suggests that, in MPVG, the values before the Last LN are counterbalanced not by the beta value but by the PE.

In conclusion, the advantage of using PE to eliminate high-value dimensions in MPVG, rather than relying on beta as in PVG, is as follows. In a Layer-wise structure, PE causes high-value dimensions to become more pronounced across dimensions. This suggests that PE counterbalances these high-value dimensions, taking over the role of LN in removing high-value dimensions. This phenomenon results in the token embedding ($\mathit{x}$) retaining values that should have been counterbalanced by PE, even after passing through the layers. Compensating for these values using PE, rather than relying solely on LN's beta, allows for more accurate counterbalancing, leading to fewer lost features compared to using LN alone to remove high-value dimensions. This suggests that, in the conventional Layer-wise structure, high-values dimensions in the Last LN were counterbalanced using only LN. However, more accurate features can be preserved by using PE to counterbalance these values.

\begin{figure*}[ht]
    \centering
    \includegraphics[width=0.7\linewidth]{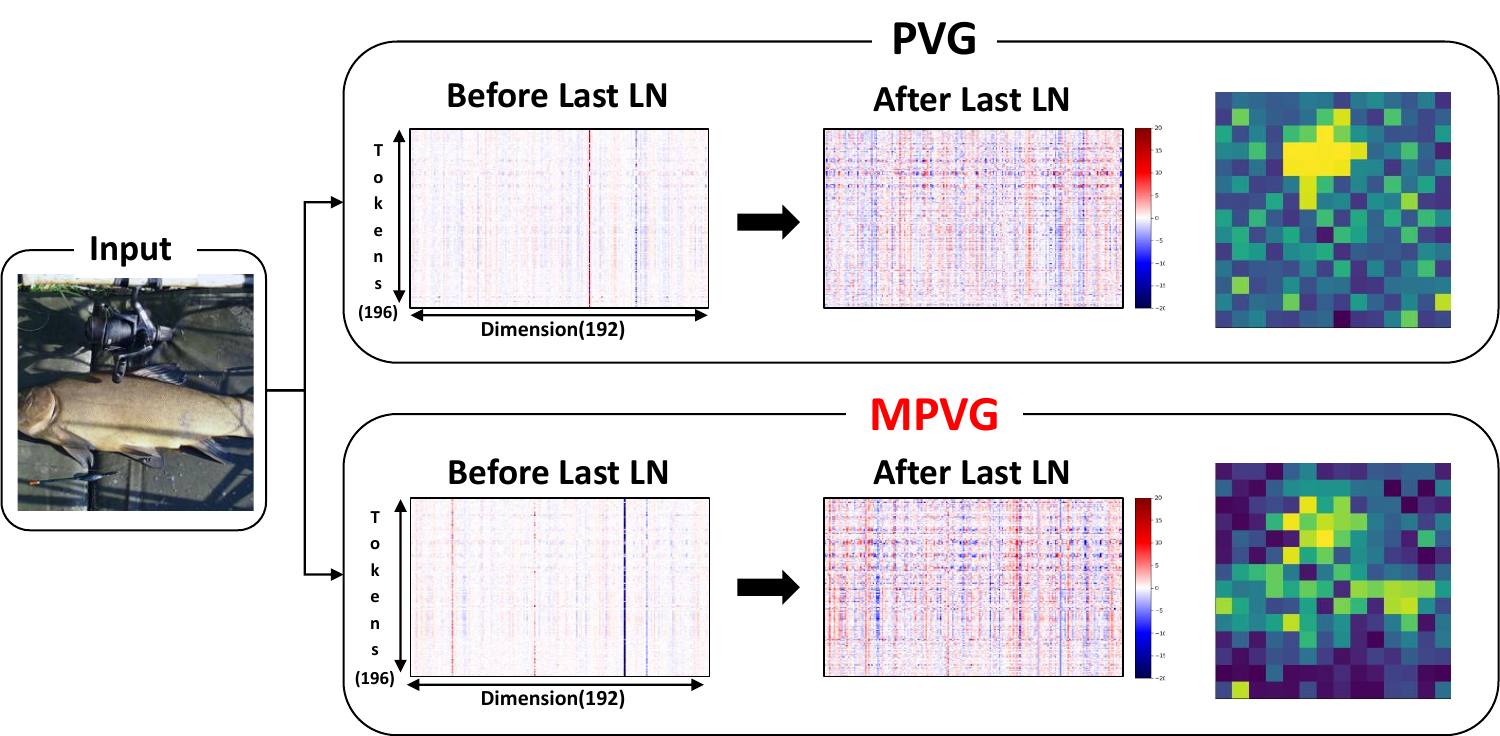}
    \caption{The visualizations and heatmaps before and after the Last LN in the PVG and MPVG methods are shown. These heatmaps represent a single sample. For the visualization heatmap, the norm values of each of the 196 tokens are calculated and visualized as heatmaps.}
    \label{supple_fig2}
\end{figure*}

As shown in Fig~\ref{supple_fig2}, MPVG captures objects more effectively than PVG, even when comparing the same samples before and after the Last LN. This demonstrates that PE in the Last LN maintains the counterbalancing directionality in a Layer-wise structure. The Last LN's role is alleviated by sustaining this directionality, enabling a richer and more accurate understanding of the objects.

\subsection{Analysis on Conflicting Results Between Class token and GAP}

In general, when the GAP method, which performs better than the class token method in image classification, is combined with the Layer-wise method, which improves the expressiveness of PE and enhances the performance of vision transformers, it leads to a decrease in performance. As we discussed, applying the GAP method in the Layer-wise structure results in a conflicting result, leading to a performance decline.

To analyze the cause, we examine the gamma and beta values of Layer Normalization (LN) in the Last LN and the cosine similarity before and after the Last LN. As shown in Figure 1, (a) represents PVG with the class token method, while (b) represents PVG with the GAP method. We observed that the variance of the beta value, an affine parameter in the Last LN, is lower in (a) compared to (b). Additionally, when observing the heatmaps before and after the Last LN, we noticed that there are no significant changes apart from the 0th row representing the class token. This suggests that the Last LN primarily focuses on the class token in the class token method.

Specifically, when comparing the cosine similarity of (a) and (b), (a) shows a value of 0.4597, while (b) shows a value of 0.2995. The high cosine similarity in (a) and the low variance of beta indicate that, in the Layer-wise structure of the class token method, the Last LN has less of a role in removing high-value dimensions compared to the GAP method. On the other hand, the lower cosine similarity and higher variance of beta in the GAP method suggest that if counterbalancing by PE does not occur in the Last LN, the Last LN alone must remove the high-value dimensions to maintain balance. This indicates that, in the Layer-wise structure, the role of PE in maintaining balance is much more critical in the GAP method, where class predictions are made directly through the average of the tokens. As shown in Figure 2, this issue can lead to less accurate results in vision transformers because the Last LN must remove high-value dimensions. In conclusion, due to the difference in the extent of the burden placed on the LN to counterbalance high-value dimensions in the class token and GAP methods, the GAP method, where the role of PE is relatively more critical, exhibits inferior performance compared to the class token method.

\subsection{Visualization on Last LN in Non-Layer-wise and Layer-wise structures.}
\begin{figure*}[ht!]
    \centering
    \includegraphics[width=1.93\columnwidth]{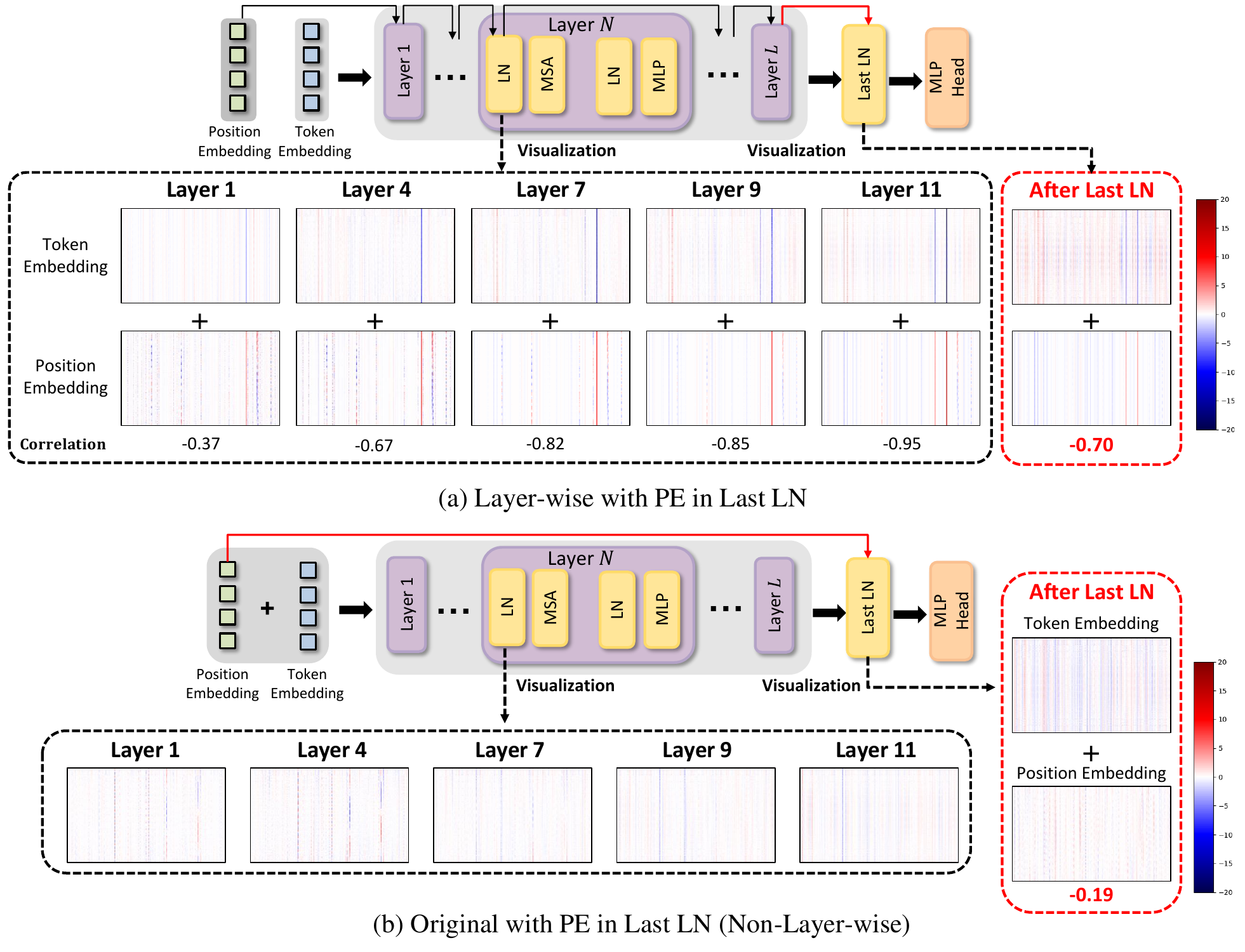}
    \caption{The difference between delivering PE to the Last LN in Non-Layer-wise and Layer-wise structures. In (a), PE is delivered to the Last LN within a Layer-wise structure, while in (b), only PE is delivered to the Last LN in a Non-Layer-wise structure.}
    \label{supple_fig3}
\end{figure*}

We provide a more detailed figure in Last LN. As shown in Fig~\ref{supple_fig3}, (a) is identical to Fig 2 in the main paper but offers a visualization after applying the Last LN. (b) visualizes Fig 5-(b) from the main paper as a heatmap. Identical to Figure 2 in main paper, the x-axis represents the dimensions, and the y-axis represents the number of tokens.

Specifically, in Fig~\ref{supple_fig3}-(b), since the structure is not Layer-wise, there is no value in the token embedding that the PE counterbalances. As a result, even if PE is added in the Last LN, there is no such directionality, leading to a decrease in performance. After applying the Last LN, the correlation values are significantly lower than (a). In contrast, in Fig~\ref{supple_fig3}-(a), because the Layer-wise structure allows PE to counterbalance the token embedding values at each layer, this directionality is maintained both before and after applying the Last LN.

\begin{table*}[ht]
\renewcommand{\arraystretch}{1.1}
\resizebox{\textwidth}{!}{
\begin{tabular}{cccccccc}
\hline
Dataset                                         & Model                         & Learning Rate & Scheduler                                                      & Weight Decay & Batch Size & Epochs                                                               & Warm-up Epochs \\ \hline
\multicolumn{1}{c|}{\multirow{5}{*}{\begin{tabular}[c]{@{}c@{}}ImageNet-1K\\ ~\cite{imagenet}\end{tabular}}}  & \multicolumn{1}{c|}{DeiT}  & 5e-4          & \begin{tabular}[c]{@{}c@{}}cosine\end{tabular} & 0.05         & 1024        & 300                                                                  & 5              \\ \cline{2-8} 
\multicolumn{1}{c|}{}                           & \multicolumn{1}{c|}{Swin}     & 1e-3          & \begin{tabular}[c]{@{}c@{}}cosine\end{tabular} & 0.05         & 1024        & 300                                                                  & 20             \\ \cline{2-8} 
\multicolumn{1}{c|}{}                           & \multicolumn{1}{c|}{CeiT}     & 5e-4          & \begin{tabular}[c]{@{}c@{}}cosine\end{tabular} & 0.05         & 1024        & 300                                                                  & 5              \\ \cline{2-8} 
\multicolumn{1}{c|}{}                           & \multicolumn{1}{c|}{T2T-ViT}  & 1e-3          & \begin{tabular}[c]{@{}c@{}}cosine\end{tabular} & 0.03         & 512        & \begin{tabular}[c]{@{}c@{}}300+10\end{tabular} & 10             \\ \hline
\multicolumn{1}{c|}{\multirow{2}{*}{\begin{tabular}[c]{@{}c@{}}CIFAR-100\\ ~\cite{cifar100}\end{tabular}}} & \multicolumn{1}{c|}{ViT-Lite} & 6e-4          & \begin{tabular}[c]{@{}c@{}}cosine\end{tabular} & 0.06         & 128        & \begin{tabular}[c]{@{}c@{}}300+10\end{tabular} & 10             \\ \cline{2-8} 
\multicolumn{1}{c|}{}                           & \multicolumn{1}{c|}{T2T-ViT}  & 5e-2          & \begin{tabular}[c]{@{}c@{}}cosine\end{tabular} & 5e-4         & 128        & 60                                                                   & -              \\ \hline
\end{tabular}}
\caption{Hyperparameter settings for image classification on ImageNet-1K and CIFAR-100.}
\label{table2-supple}
\end{table*}

\begin{table*}[t]
\renewcommand{\arraystretch}{1.1}
\resizebox{\textwidth}{!}{
\begin{tabular}{cccccccccc}
\hline
Dataset                                                        & Model          & Framework  & \begin{tabular}[c]{@{}c@{}}Backbone\\ Prei-train\end{tabular} & \begin{tabular}[c]{@{}c@{}}Crop \\ Size\end{tabular} & Optimizer & LR   & Scheduler & Weight Decay & Batch Size \\ \hline
\begin{tabular}[c]{@{}c@{}}COCO 2017\\~\cite{coco}\end{tabular} & ViT-Adapter-Ti & Mask R-CNN & DeiT-Ti                                                       & 1024                                                 & AdamW     & 1e-4 & 3x+MS     & 0.05         & 8          \\ \hline
\end{tabular}}
\caption{Hyperparameter settings for object detection on COCO 2017.}
\label{table3-supple}
\end{table*}

\begin{table*}[t!]
\renewcommand{\arraystretch}{1.1}
\resizebox{\textwidth}{!}{
\begin{tabular}{cccccccccc}
\hline
Dataset & Model          & Framework & \begin{tabular}[c]{@{}c@{}}Backbone\\ Prei-train\end{tabular} & \begin{tabular}[c]{@{}c@{}}Crop \\ Size\end{tabular} & Optimizer & LR    & Scheduler & Weight Decay & Batch Size \\ \hline
\begin{tabular}[c]{@{}c@{}}ADE20K\\~\cite{ade20k}\end{tabular}  & ViT-Adapter-Ti & UperNet   & DeiT-Ti                                                       & 512                                                  & AdamW     & 12e-5 & 160K      & 0.01         & 8          \\ \hline
\end{tabular}}
\caption{Hyperparameter settings for semantic segmentation on ADE20K.}
\label{table4-supple}
\end{table*}

\end{document}